\documentclass{article}

\usepackage[final,nonatbib]{neurips_2021}
\usepackage{graphicx}
\usepackage{caption}
\usepackage{subcaption}

\usepackage[utf8]{inputenc} 
\usepackage[T1]{fontenc}    
\usepackage{hyperref}       
\usepackage{url}            
\usepackage{booktabs}       
\usepackage{amsfonts}       
\usepackage{nicefrac}       
\usepackage{microtype}      
\usepackage{xcolor}         
\usepackage[square,numbers]{natbib}
\usepackage{authblk}
\usepackage{wrapfig,lipsum,booktabs}

\title{Deep Learning Methods for Daily Wildfire Danger Forecasting}

    \makeatletter
\def\@fnsymbol#1{\ensuremath{\ifcase#1\or \dagger\or \ddagger\or
   \mathsection\or \mathparagraph\or \|\or **\or \dagger\dagger
   \or \ddagger\ddagger \else\@ctrerr\fi}}
    \makeatother
   
\author[1,2,*]{\underline{Ioannis Prapas}}
\author[1,2,*]{\underline{Spyros Kondylatos}}
\author[1]{Ioannis Papoutsis}
\author[2]{Gustau Camps-Valls}
\author[2]{Michele Ronco}
\author[2]{Miguel-Ángel Fernández-Torres}
\author[2]{Maria Piles Guillem}
\author[3]{Nuno Carvalhais}
\affil[1]{IAASARS\thanks{Institute for Astronomy, Astrophysics, Space Applications and Remote Sensing} , National Observatory of Athens}
\affil[2]{Image Processing Laboratory (IPL), Universitat de València}  
\affil[3]{Max Planck Institute for Biogeochemistry}
\affil[*]{Authors contributed equally}

\begin{document}
\setlength{\belowcaptionskip}{-10pt}
\maketitle

\begin{abstract}
 
Wildfire forecasting is of paramount importance for disaster risk reduction and environmental sustainability. We approach daily fire danger prediction as a machine learning task, using historical Earth observation data from the last decade to predict next-day's fire danger. To that end, we collect, pre-process and harmonize an open-access datacube, featuring a set of covariates that jointly affect the fire occurrence and spread, such as weather conditions, satellite-derived products, topography features and variables related to human activity. We implement a variety of Deep Learning (DL) models to capture the spatial, temporal or spatio-temporal context and compare them against a Random Forest (RF) baseline. We find that either spatial or temporal context is enough to surpass the RF, while a ConvLSTM that exploits the spatio-temporal context performs best with a test Area Under the Receiver Operating Characteristic of 0.926. Our DL-based proof-of-concept provides national-scale daily fire danger maps at a much higher spatial resolution than existing operational solutions.


\end{abstract}

\section{Introduction and Application Context}

\setcounter{footnote}{0} 

Globally, wildfires are a major natural hazard, which disrupts natural ecosystem services, causes loss of lives, properties and infrastructure \cite{pettinari_fire_2020}, and contributes to carbon dioxide emissions.
Climate change is playing an increasing role in determining wildfire regimes and is expected to aggravate wildfires in most parts of the Earth, 
and quite prominently in the wider Mediterranean~\cite{turco_exacerbated_2018}. Thus, assessing the likelihood of large fire events is of utmost importance for fire management services.

Traditionally, Fire Danger is predicted by state-wide Fire Danger Rating systems (FDRS), such as the Canadian FDRS \footnote{\href{https://cwfis.cfs.nrcan.gc.ca/background/summary/fdr}{https://cwfis.cfs.nrcan.gc.ca/background/summary/fdr}} or the Australian FDRS \footnote{\href{https://www.afac.com.au/initiative/afdrs}{https://www.afac.com.au/initiative/afdrs}}. In Europe, the European Forest Fire Information system (EFFIS) \cite{SANMIGUELAYANZ201319} uses a similar approach to the Canadian FDRS, which is based on a meteorologically derived Fire Weather Index (FWI).
Our work considers vegetation characteristics captured from time-series of Earth Observation (EO) data, and information layers connected to human activity, in addition to meteorological data, and learns to predict fire danger in a data-driven way.

We create a carefully harmonized datacube that contains both historical wildfires and covariates that drive these wildfires. On top, we train Deep Learning (DL) models to forecast the next day's fire danger and evaluate these models in real conditions. This proof-of-concept is developed in partnership with Fire Brigade officials, towards an operational national fire management support tool, for effectively allocating firefighting resources, designing and applying mitigation measures and timely responding to wildfire disasters. 


\section{Related Work and Contributions}



Wildfire danger forecasting is not a typical Machine Learning (ML) task and poses three great challenges. First, wildfires are caused by the complex interactions of the fire drivers (weather conditions, land and vegetation characteristics, human activities) \cite{hantson_status_2016}. 
Second, wildfire occurrence is inherently stochastic; the same Earth system conditions may lead to major fire events or not. This makes sampling negative examples complex, as the lack of a fire event does not mean lack of fire danger. Third, wildfires are rare events, which can potentially lead to a highly imbalanced dataset. 

\citet{reichstein_deep_2019} highlighted  fire occurrence and spread as typical examples of applications where both spatial and temporal context are highly relevant and could profit from DL modeling. However, relevant DL solutions remain still in their infancy \cite{zhang_forest_2019,huot_deep_2020}.
\citet{huot_deep_2020} propose a modern DL approach for fire likelihood estimation, viewing it as a segmentation task and employing U-NET 
type architectures to predict the daily or weekly-aggregated fire masks of MODIS \cite{giglio_collection_2016}. However, they make binary predictions (fire, no-fire) which do not account for the stochasticity of fire occurrence, and do not identify the fire ignitions associated with the fire masks, which can lead to data leakage, predicting fires that are already burning. \citet{zhang_forest_2019} combine weather and remote sensing data as input to a Convolutional Neural Network (CNN) to model forest fire susceptibility. However, they only exploit the spatial context, ignoring the temporal aspect of the phenomenon. Moreover, a common problem in both works is that they define random training-testing splits, which overestimates the models' performance in the real world. Instead, fire prediction should be treated as a forecasting task \cite{oliveira2021evaluation}, that is to train the models in the past to predict the future. Other related work \cite{jain_review_2020}  either focuses on a small region or uses shallow models that cannot 
handle the spatio-temporal context.

Our work builds upon the presented methods and moves a step forward on better estimating fire likelihood, making the following contributions:
    \textbf{First, we create, harmonize, curate and openly publish a large datacube} \cite{prapas_datacube_2021} containing fire driver variables and burned areas for the period 2009-2020 in Greece, a fire-prone country of the Mediterranean.
    \textbf{Second, we rigorously formulate fire danger forecasting as an ML problem.} For that purpose, we identify the fire events associated with large burned areas, along with their starting date, which allows us to model the joint probability of fires occurring and becoming large. Then, we treat fire prediction as a forecasting task by doing a time split to create the training, validation and test sets. Moreover, we do a careful negative sampling in order to account for the dataset imbalance.
    \textbf{Third, we propose and compare a variety of DL models} that are able to capture spatial, temporal or spatio-temporal context. We present an operational prototype, which predicts  the next day's fire danger at a national scale in Greece.



\section{Data harmonization}\label{sec:dataset}
Our dataset covers most of Greece and surrounding area, and contains 
climate, vegetation and human activity variables, which influence the ignition and spread of a fire \cite{pettinari_fire_2020}. The target, as explained in Section \ref{sec:setup}, is defined by the burned area resulting from a fire that started on a given day. We gather all variables 
and harmonize them into a $1$ km $\times$ $1$ km $\times$ $1$ day resolution datacube, which covers an area of $700$ km $\times$ $562$ km for the years 2009-2020.

More specifically, we gather the following data: 1. \textbf{Daily weather data} from ERA-5 Land \cite{munoz2021era5}, for 5 hours (04:00, 08:00, 12:00, 16:00, 20:00) of 2 m temperature, 10 m wind $u$-component, 10 m wind $v$-component and total precipitation, available in 9 km spatial resolution. 2. \textbf{Satellite variables} from MODIS downloaded from Nasa's portal \footnote{\href{https://modis.gsfc.nasa.gov/data/}{https://modis.gsfc.nasa.gov/data/}}, 
    including Leaf Area Index (LAI) and Fraction of Photosynthetically Active Radiation (Fpar), available in a 8-daily temporal and 500 m spatial resolution; Normalized Difference Vegetation Index (NDVI) and Enhanced Vegetation Index (EVI), available in a 16-daily temporal and 500 m spatial resolution; and Day/Night Land Surface Temperature (LST), available in a daily temporal and 1 km spatial resolution. 3. \textbf{Roads density.} Road networks are extracted from OpenStreetMap\footnote{\href{https://www.openstreetmap.org/}{https://www.openstreetmap.org/}}, in order to calculate street density at 1 km spatial resolution. 4. \textbf{Population density} extracted from WorldPop, where yearly measurements are available. 5. \textbf{Land cover} extracted from Copernicus Corine Land Cover (CLC) \cite{buttner_corine_2014} for years 2006, 2012 and 2018, at 100 m spatial resolution. 6. \textbf{Topography variables} from Copernicus EU-DEM \cite{bashfield_continent-wide_2011}
    including elevation, aspect and slope at 30 m spatial resolution. 7. \textbf{Historical burned areas} from the European Forest Fire Information System (EFFIS) \cite{SANMIGUELAYANZ201319} 
    containing burned areas larger than 30 hectares (ha). The EFFIS burned areas are intersected with the MODIS active fires product \cite{giglio_collection_2016} to recover the start date of the fire.

The data sources are aggregated into a datacube of 19 features. 
The dynamic attributes (13 features) are the daily min/max 2 m temperature, min/max $u$-component and $v$-component of wind, max total precipitation, Fpar, LAI, Day/Night LST, NDVI and EVI. To simulate a real-time setting, the satellite attributes are shifted forward one day, as MODIS provides near real-time data at a daily temporal resolution and a day's observations are available after the satellite passes the location. The weather attributes are not shifted as we assume them to be available from forecasts. The static attributes (6 features) are the CLC, mean elevation, aspect and slope, along with roads and population density. We \textbf{publish this datacube} in order to be reused by the community \cite{prapas_datacube_2021}.

\section{Experiments}\label{sec:experiments}
\subsection{Experimental setup}\label{sec:setup}

\begin{figure}[b]
 \centering
  \includegraphics[width=\linewidth]{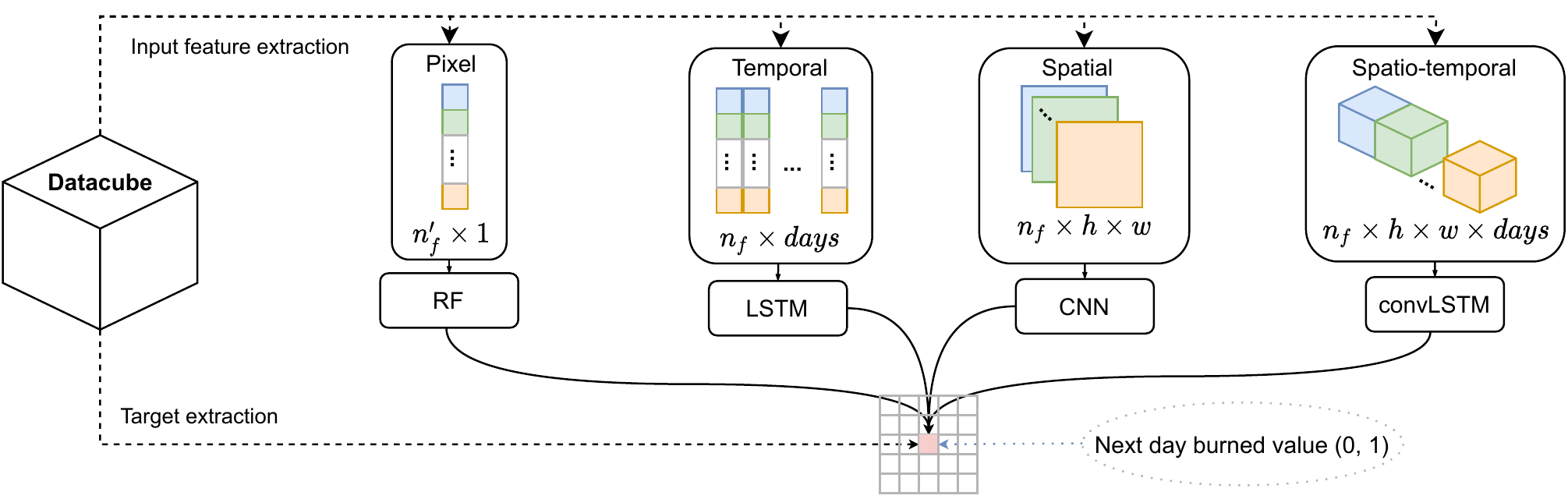}

 \caption{Four different types of dataset samples (pixel, temporal, spatial, and spatio-temporal) are extracted from the datacube and then fed to the corresponding models (RF, LSTM, CNN, and convLSTM). For a dataset sample, $n_f$ is the number of input features, $days$  is the number of days in the time series, $h$ is the height and $w$ is the width, wherever applicable. The pixel dataset has a different number of input features ($n_f'$), because it also includes aggregations of the dynamic features.}
 \label{fig:dataset-extraction}
\end{figure}

Wildfire emergency response decision makers are primarily interested in the next day spatial probability of large wildfires~\cite{https://doi.org/10.1002/env.2269}. 
If $FI$ is the event that a wildfire ignites and $FL$ is the event that a wildfire becomes large (> 30 ha), we wish to predict the joint probability of a fire occurring and becoming a large fire $FI \cap FL$~\cite{https://doi.org/10.1002/env.2269}. This translates to $P(FI \cap FL) = P(FI)P(FL|FI)$.


To model $P(FI \cap FL)$, we extract our training, validation and testing sets from the datacube introduced above, as follows: 
    
    For a given pixel (i.e. a cell representing a $1$ km  $\times$ $1$ km square region) and a given day, we extract input-target pairs for $4$ different modeling modalities.  First, the \textit{pixel dataset}, where we extract the input attributes and their last $10$-days average (only for the dynamic input attributes). Second, the \textit{temporal dataset}, where we extract the last $10$-days time-series of the input attributes. Third, the \textit{spatial dataset}, where we extract $25$ km $\times$ $25$ km patches centered spatially around the given pixel. Fourth, the \textit{spatio-temporal dataset}, where we extract $25$ km  $\times$ $25$ km $\times$ $10$ days blocks centered spatially around the given pixel. Static variables are repeated in the time dimension. All input variables are used, except for the CLC, because it is used to determine the negative samples, as discussed next. \textit{The target is for all types of datasets the same}; the next-day burned value of the given pixel. This dataset extraction process is visualized in Figure \ref{fig:dataset-extraction}, where each type of dataset is then fed to a different type of ML model. 
    
    Positive examples (pixels contained in burned areas) are all included in the datasets. We sample two times more negative examples (pixels not contained in burned areas), extracted from days when no fire occurred.
    Furthermore, we stratify the negative sampling to follow the land cover distribution of the positive examples (burned pixels). This stratification strategy makes the distinction between positives and negatives harder for the models, and prevents them from learning trivial mappings.
    
    We do a time split using the years 2009-2018 for training, 2019 for validation and 2020 for testing. All the datasets consist of 
15731 training (10380 non-fire, 5351 fire),  1337 validation (1075 non-fire, 262 fire), and 1286 testing (975 non-fire, 311 fire) examples. 

%
%
%

\subsection{Models and training}\label{sec:models}

We train a different model for each type of dataset introduced above, as depicted in Figure \ref{fig:dataset-extraction}. 

For the pixel dataset, we consider a Random Forest (RF) baseline model, which has been extensively used in related work~\cite{jain_review_2020}. Its hyperparameters are chosen based on the validation Area Under the Receiver Operating Characteristic (AUROC). The final RF consists of 100 trees, with no max depth set, the minimum number of samples required to split is 2 and the minimum number of samples required to be at a leaf node is 1. 

For the temporal, spatial and spatio-temporal datasets, typical large DL models quickly overfit, because our datasets are relatively small. Thus, we develop some custom models, whose architectures and hyperparameters have been chosen based on the validation loss. All linear layers, but the last, are followed by a dropout with probability $p=0.5$ and the ReLU activation function. The data is standardized before serving as input. 
The DL models are trained for 50 epochs with the binary cross-entropy loss with $\ell_2$-norm regularization, the Adam optimizer and a batch size of $128$.

For the temporal dataset, we use a Long-Short Term Memory (LSTM) 
architecture that is able to capture temporal dynamics. An LSTM layer with $64$ neurons is followed by two linear hidden layers with $64$ and $32$ neurons, respectively, and an output $2$-class softmax layer. The model is trained with a $0.01$ weight decay and $0.001$ learning rate. 

For the spatial dataset, we use a CNN 
architecture, known to effectively capture spatial features. A CNN layer with $16$ filters with $3\times 3$ kernels, padding of $1$, and stride of $1$ is followed by a $2\times 2$ max-pooling layer. This is followed by a dropout and two linear layers with $16$ and $8$ neurons, respectively, before the final 2-class softmax layer. The model is trained with a $0.03$ weight decay and $0.0004$ learning rate.

For the spatio-temporal dataset, we use a ConvLSTM \cite{shi_convolutional_2015}, which is known to capture spatio-temporal features of the input, combining the strengths of the LSTM and CNN components. A ConvLSTM layer with $16$ filters with $3\times 3$ kernels is used, followed by an architecture similar to the aforementioned CNN. The model is trained with a $0.03$ weight decay and $0.0001$ learning rate.


\subsection{Results}

\begin{wraptable}{r}{8.5cm}
 \caption{Performance of the models on the test set (year 2020)}

\label{table1}
\centering
\begin{tabular}{|c | c c c  c|} 
    \toprule
 Model & Precision & Recall & $F_1$ & AUROC \\
    \midrule
 RF & \textbf{0.832} & 0.508 & 0.631 & 0.898 \\ 
 LSTM & 0.741 & \textbf{0.762} & \textbf{0.751} & 0.920 \\
 CNN & 0.732 & 0.553 & 0.63 & 0.910 \\
 ConvLSTM & 0.798 & 0.646 & 0.714 & \textbf{0.926} \\
 \bottomrule

\end{tabular}
\end{wraptable} 

The evaluation is performed over the test set, which contains all the fire events during the year 2020. In Table~\ref{table1}, we report the precision, recall and $F_1$ score for the positive class (burned), together with the AUROC. 

While RF achieves the best precision, it has the lowest recall and $F_1$ score, together with the CNN, i.e. they both miss fire events. The LSTM and ConvLSTM perform better regarding these metrics. However, since we want to output the fire likelihood and not just a hard binary prediction, we consider the AUROC to be the most important metric for the evaluation. In terms of AUROC, the RF achieves the weakest performance, while ConvLSTM achieves the best one, suggesting that the best performance can be achieved when both spatial and temporal contexts are combined. Moreover, whilst all DL models outperform RF, LSTM and ConvLSTM have better AUROC than CNN, showing that temporal context is more important than spatial context for wildfire danger forecasting.

%

\begin{figure}
\centering

\begin{subfigure}[t]{0.205\linewidth}
\centering
RF
    \includegraphics[width=1.0\textwidth, trim=5cm 17cm 32.2cm 4cm,clip]{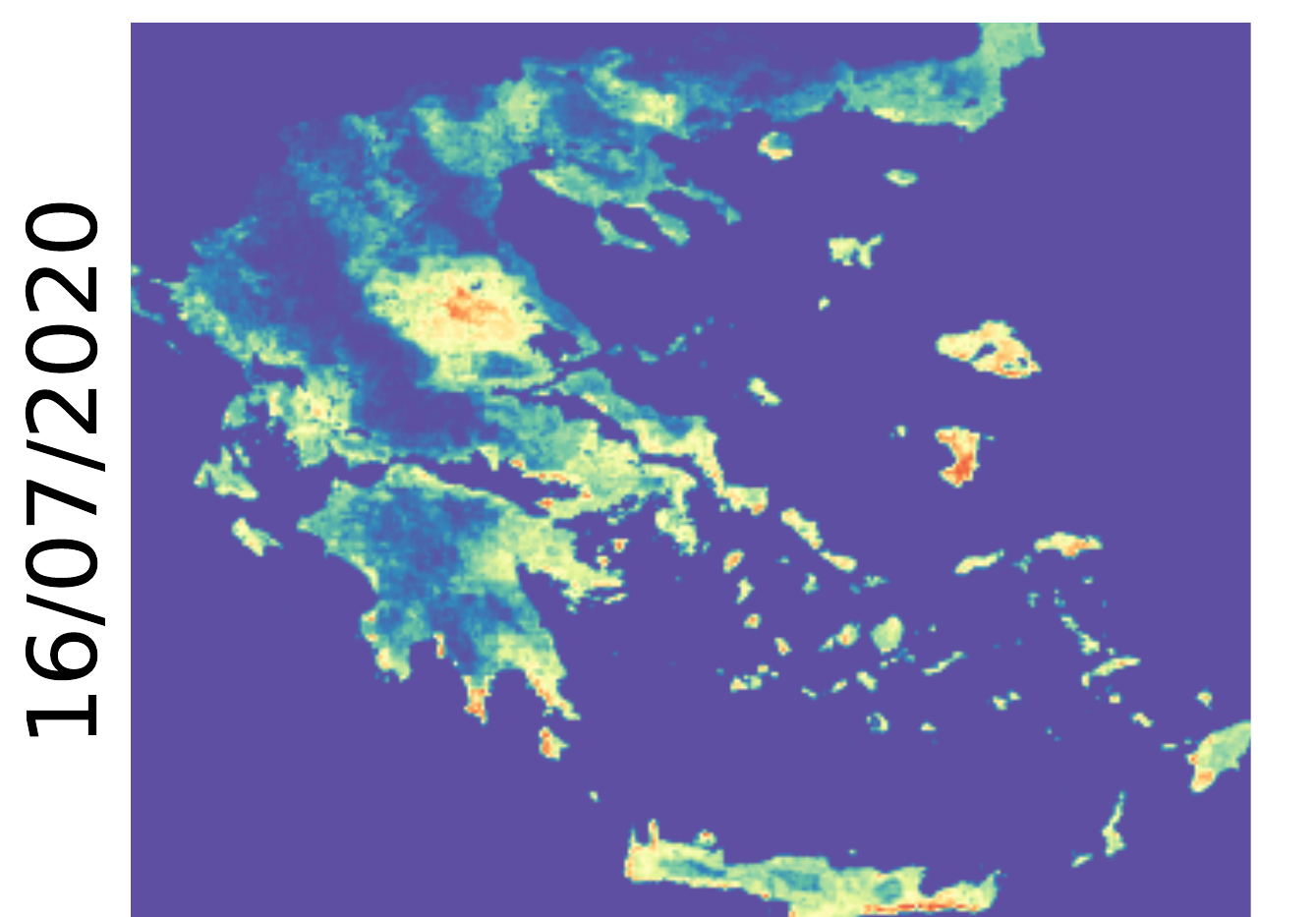}
\label{fig:rf_1}
\end{subfigure}
\begin{subfigure}[t]{0.19\textwidth}
    \centering
    LSTM
  \includegraphics[width=0.92\textwidth, trim=6cm 17cm 33cm 4cm,clip]{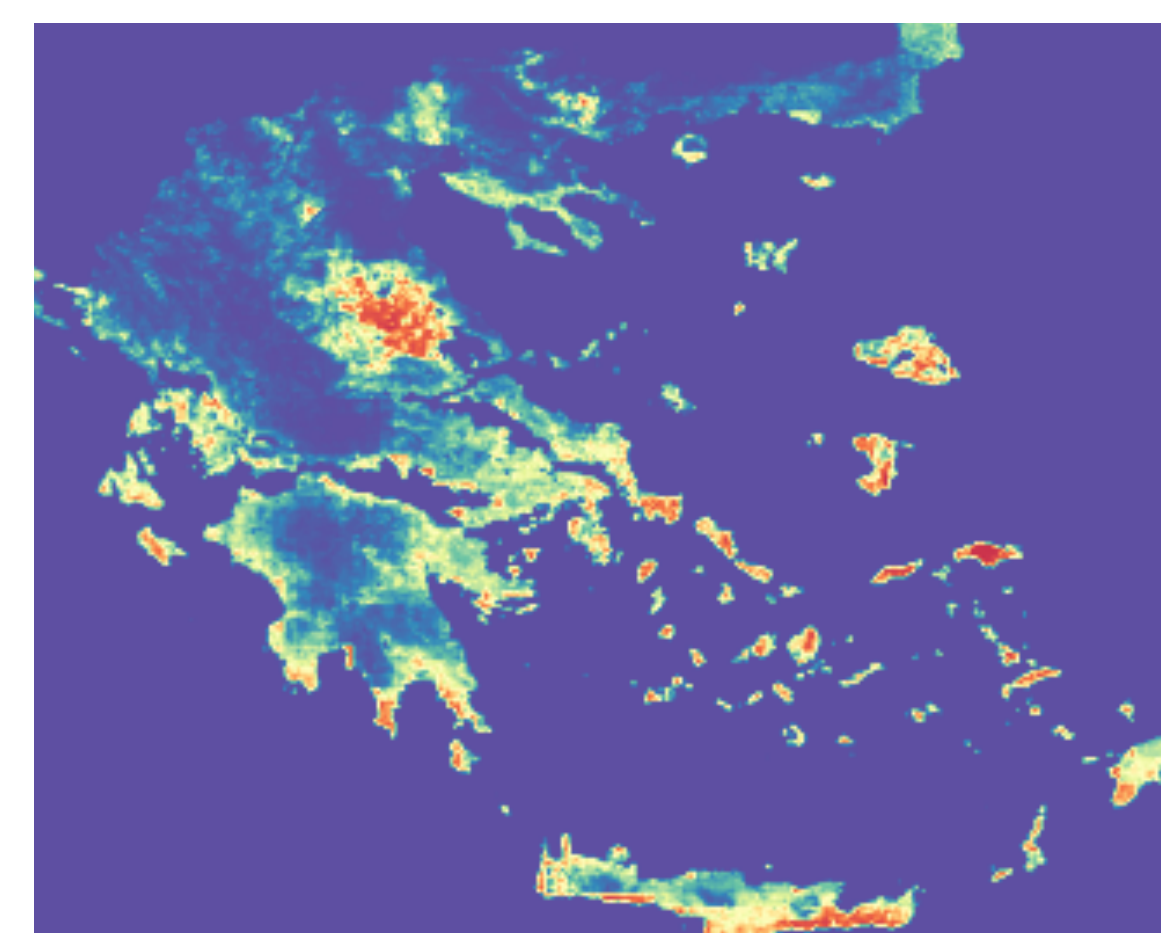}
\label{fig:lstm_1}
\end{subfigure}
\begin{subfigure}[t]{0.19\textwidth}
    \centering
    CNN
    \includegraphics[width=0.92\textwidth, trim=6cm 17cm 33cm 4cm,clip]{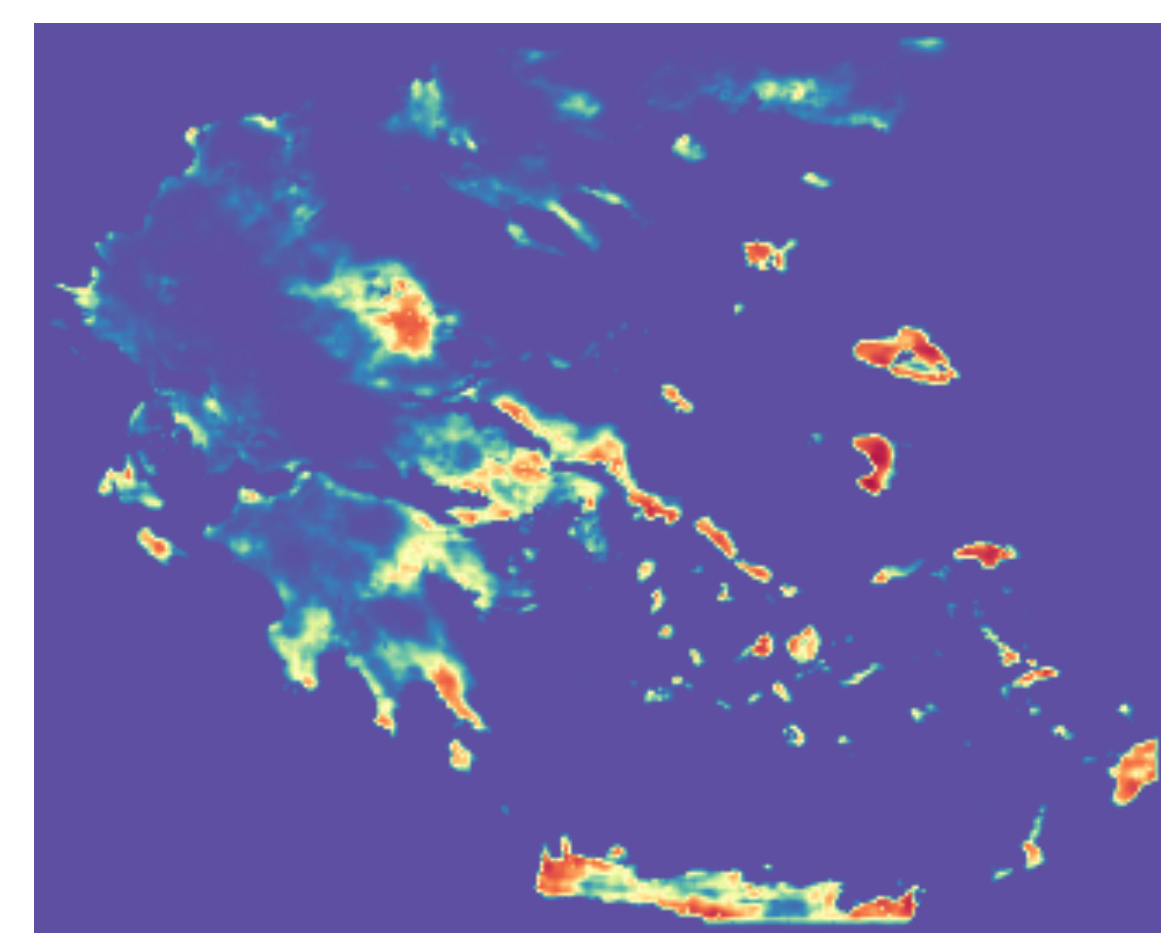}
\label{fig:cnn_1}
\end{subfigure}
\begin{subfigure}[t]{0.19\textwidth}
    \centering
    ConvLSTM
    \includegraphics[width=0.92\textwidth, trim=6cm 17cm 33cm 4cm,clip]{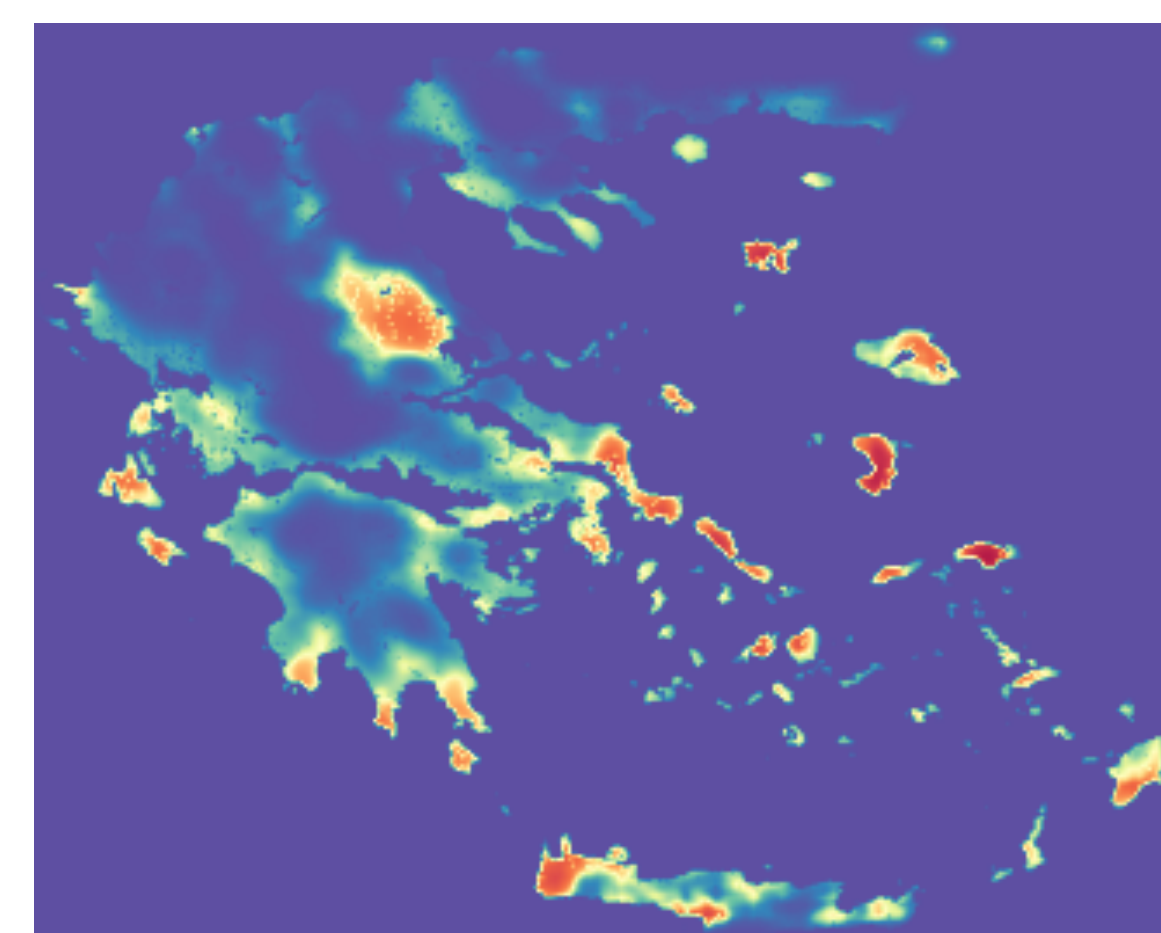}
\label{fig:clstm_1}
\end{subfigure}
\begin{subfigure}[t]{0.19\textwidth}
    \centering
    FWI
    \includegraphics[width=0.92\textwidth, trim=6cm 17cm 33cm 4cm,clip]{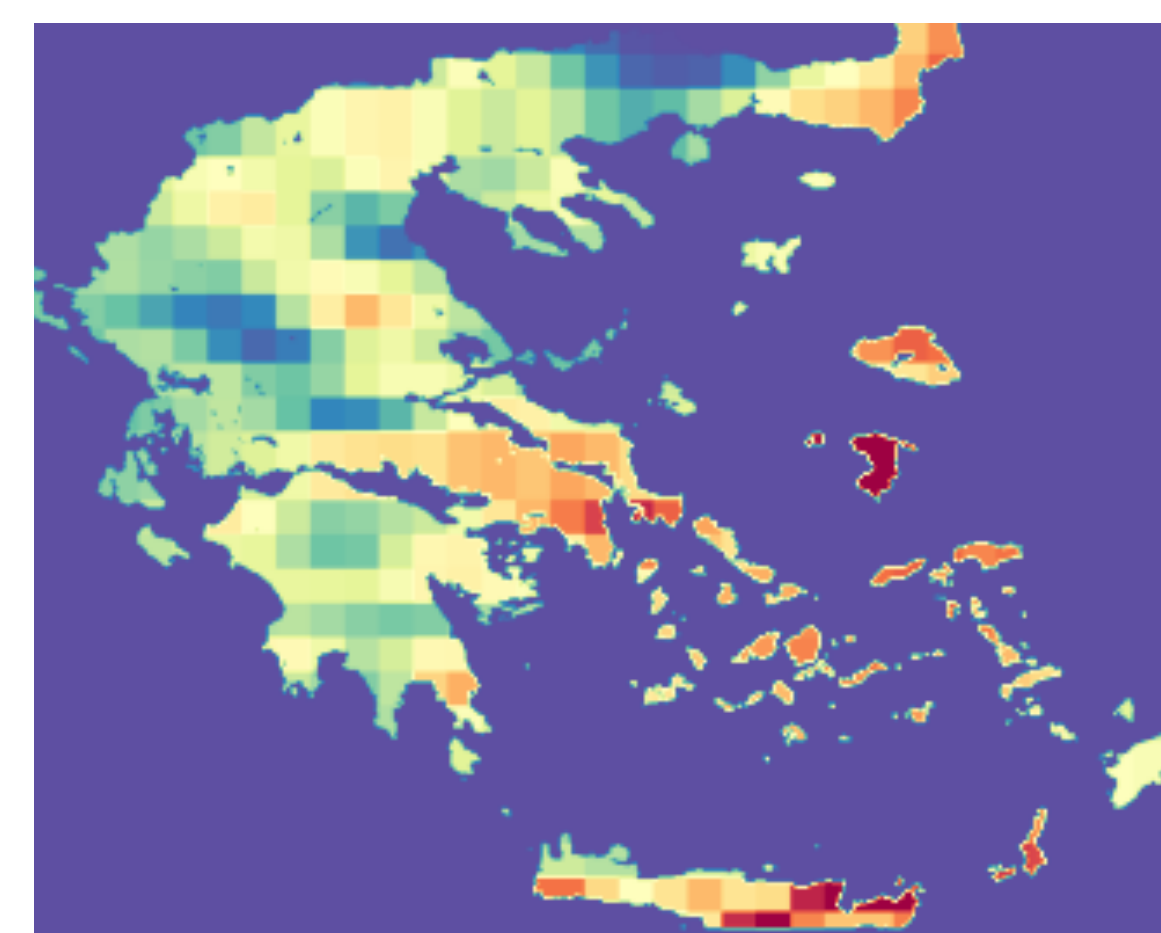}
\label{fig:fwi_1}
\end{subfigure}

\begin{subfigure}[t]{0.205\textwidth}
\centering
    \includegraphics[width=1.0\textwidth, trim=5cm 17cm 32.2cm 4cm,clip]{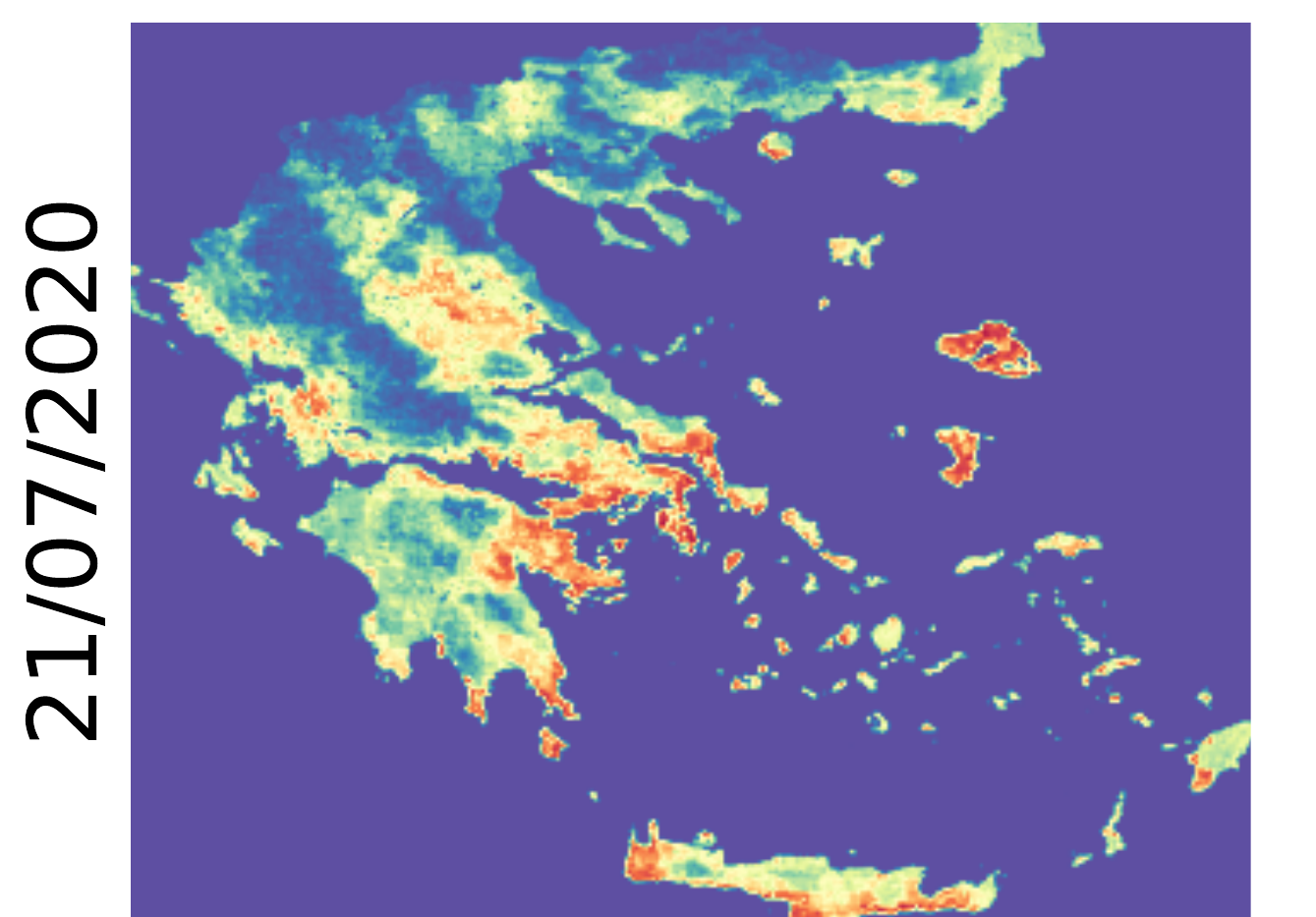}
\label{fig:rf_2}
\end{subfigure}
\begin{subfigure}[t]{0.19\textwidth}
\centering
  \includegraphics[width=0.92\textwidth, trim=6cm 17cm 33cm 4cm,clip]{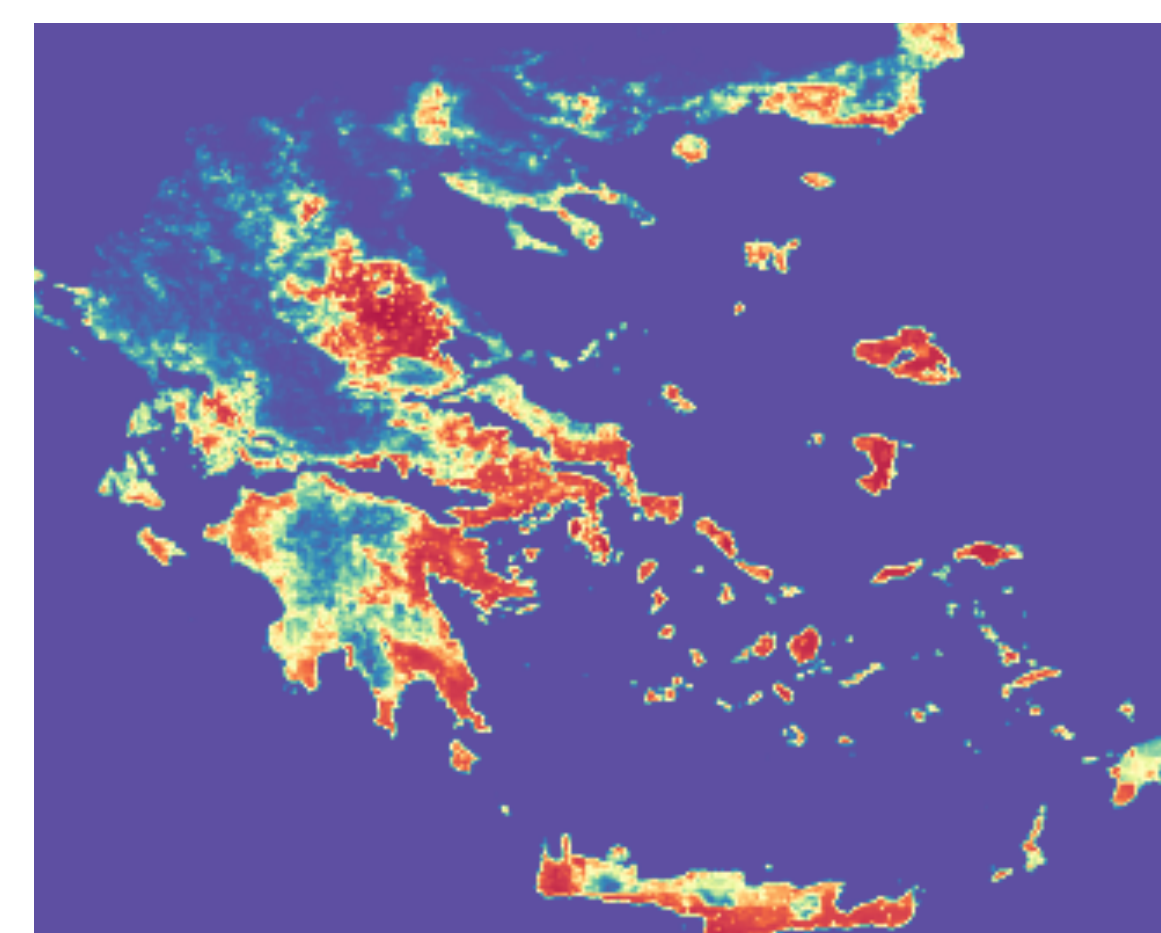}
\label{fig:lstm_2}
\end{subfigure}
\begin{subfigure}[t]{0.19\textwidth}
\centering
    \includegraphics[width=0.92\textwidth, trim=6cm 17cm 33cm 4cm,clip]{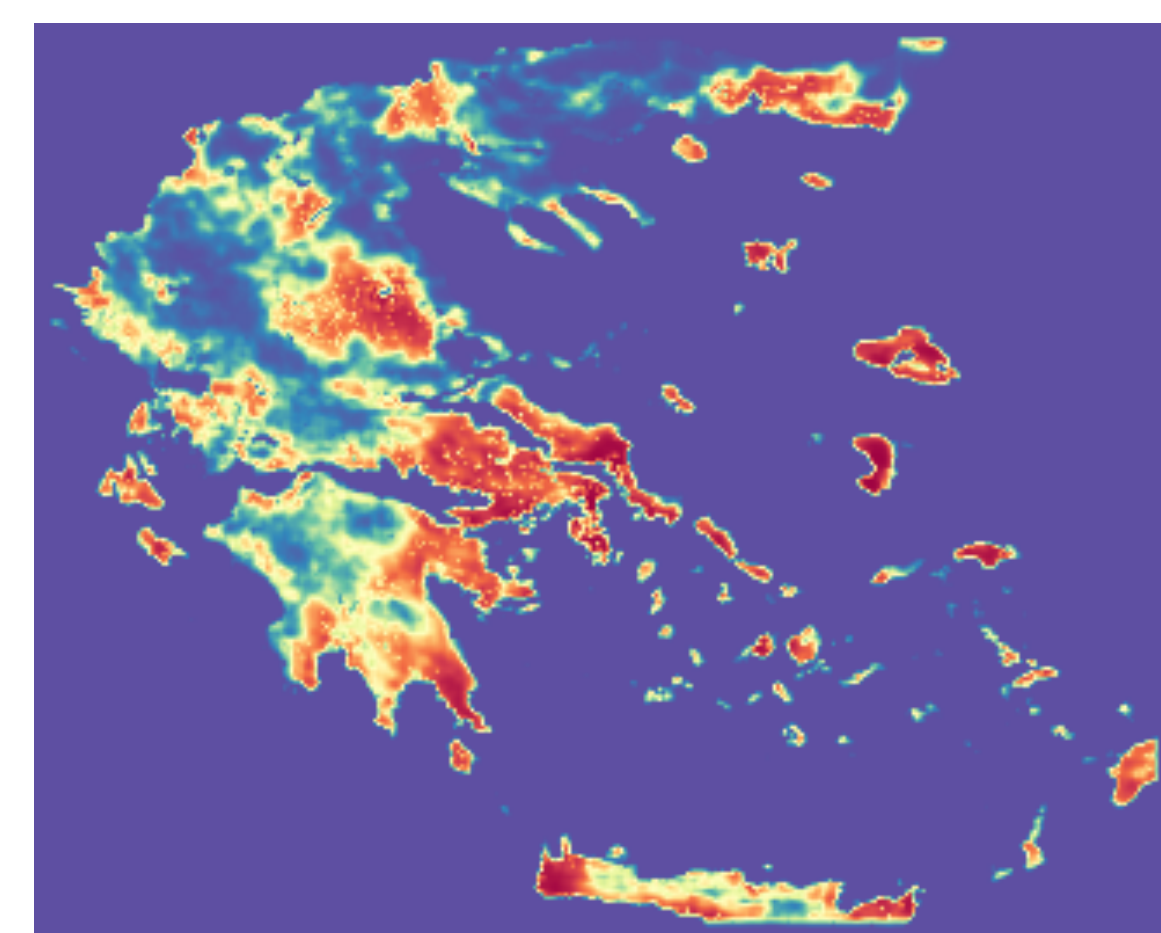}
\label{fig:cnn_2}
\end{subfigure}
\begin{subfigure}[t]{0.19\textwidth}
\centering
    \includegraphics[width=0.92\textwidth, trim=6cm 17cm 33cm 4cm,clip]{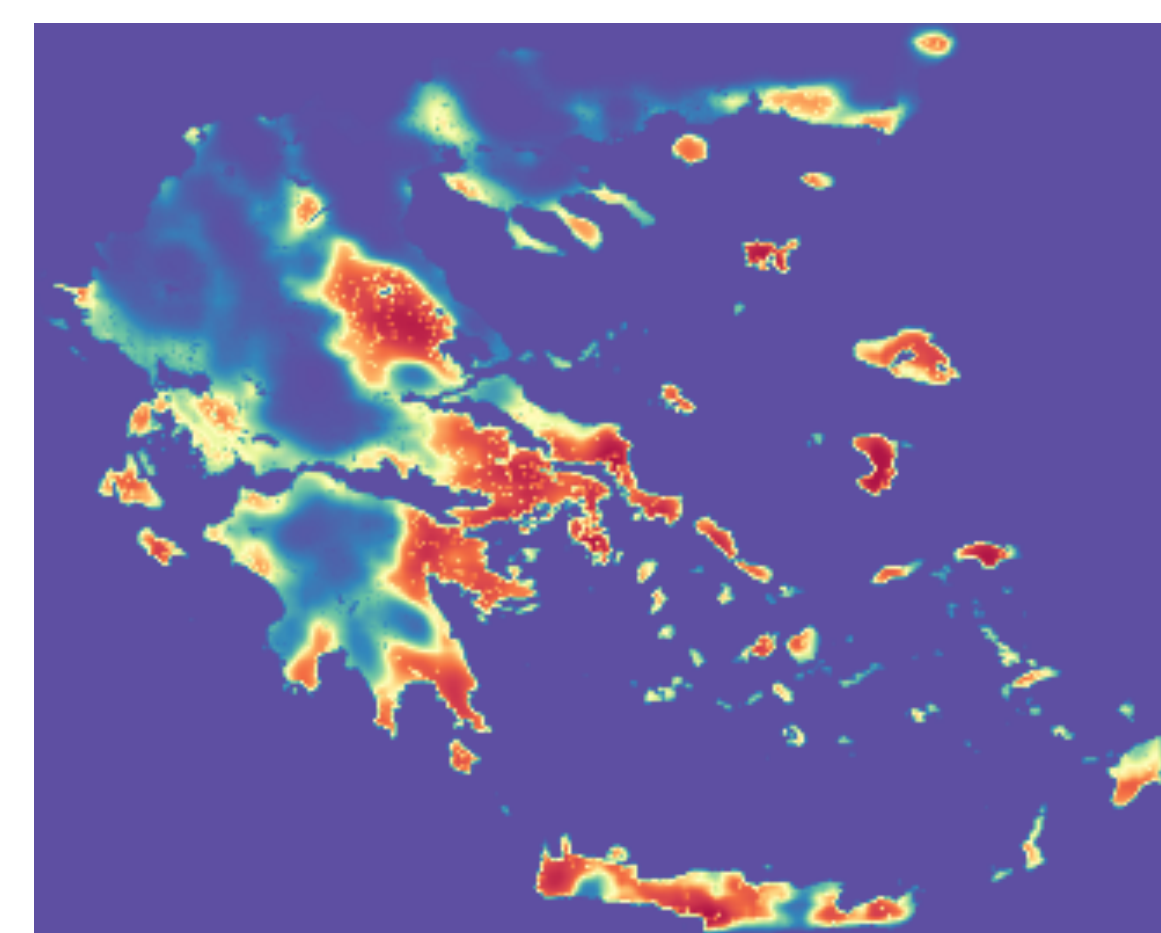}
\label{fig:clstm_2}
\end{subfigure}
\begin{subfigure}[t]{0.19\textwidth}
\centering
    \includegraphics[width=0.92\textwidth, trim=6cm 17cm 33cm 4cm,clip]{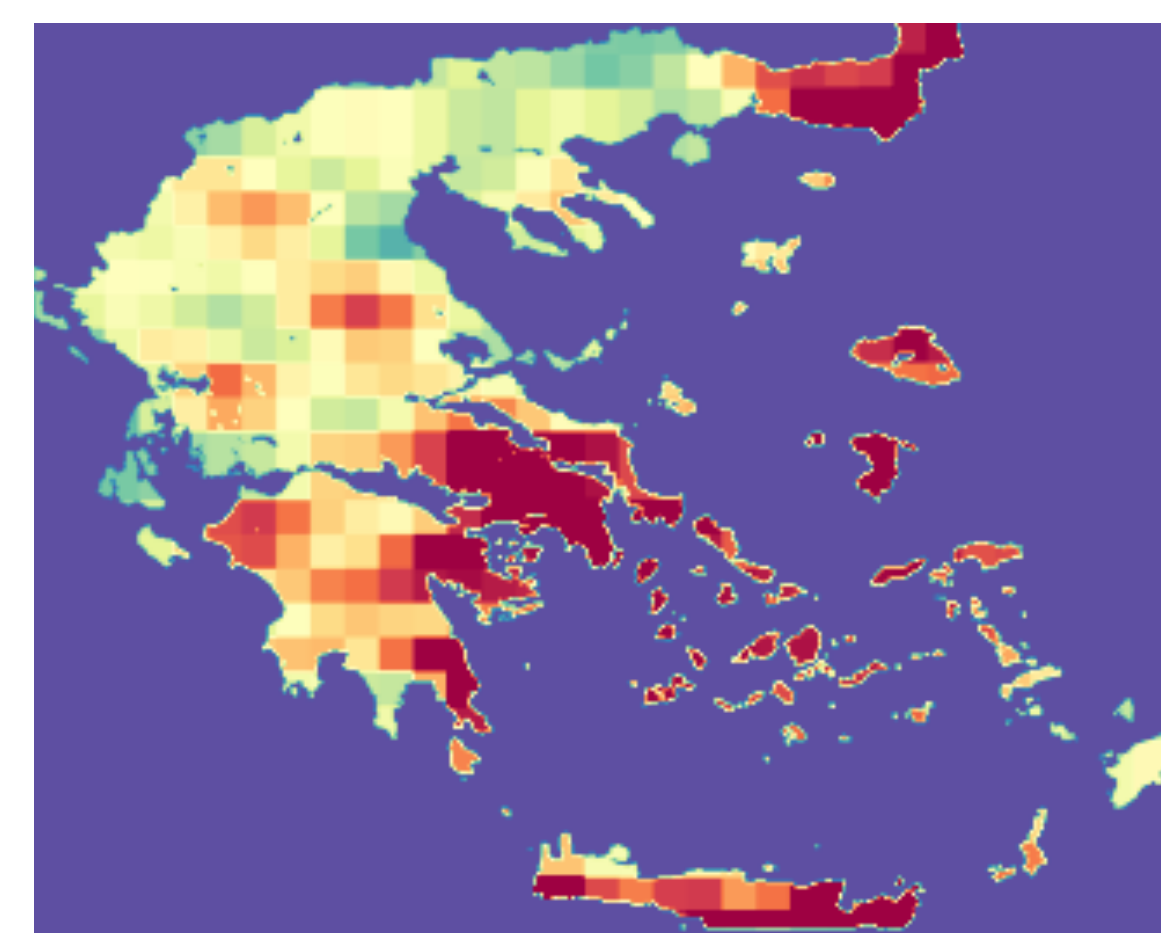}
\label{fig:fwi_2}
\end{subfigure}


\caption{Maps produced by different models and the day's FWI for two dates with medium (top) and high (bottom) fire danger. Extreme danger is shown in dark red and no danger in blue.}
\label{fig:fire danger}
\end{figure}

However, a single metric like AUROC does not tell the whole story. In Figure \ref{fig:fire danger} we visualize all models' predictions for two separate days in 2020. Next to them, we plot the  FWI, which is commonly used operationally as an index related to fire danger and is provided at a $8$ km $\times$ $8$ km spatial resolution. At a first glance, all the models seem to follow the same wildfire danger patterns, with some subtle differences, yet all of them provide finer details and more realistic susceptibility patterns compared to FWI. The RF map outputs generally lower probabilities than the other models. 
Lastly, CNN and ConvLSTM maps are visually smoother, as they exploit the spatial context which is overlapping in neighboring cells.


\section{Discussion}\label{sec:discussion}

The results on the reported metrics for daily fire labels (Table~\ref{table1}), demonstrate the DL models' potential to estimate wildfire danger. While, in terms of AUROC we see the advantage of DL models, when looking at the other metrics there is not a clear winner. However, the precision and recall values can be tuned by choosing a class weight for the positive class or by sampling more or less negative examples. Visually, the generated national-scale daily prediction maps (Figure~\ref{fig:fire danger}) have structure, contain both large and local scale patterns, and show spatio-temporal variability.

Further research will scale to the greater fire-prone Mediterranean area. This will allow to test how well our models generalize to similar biomes. It will also add data, which can allow us to improve with deeper and more sophisticated DL architectures to account for richer temporal information to detect legacy effects, as well as long-range spatial connections.






While DL models are able to make predictions with high accuracy, their deterministic nature prevents them from providing uncertainty estimates for their predictions. Wildfire occurrence, however, presents a high degree of stochasticity, therefore uncertainty assessment is critical in decision making. Hence, Bayesian ML, DL \cite{jospin_hands-bayesian_2020} approaches offer a promising route in that direction. 
 



Given that wildfires are a phenomenon with huge impact on human lives and the ecosystem, explaining can be more important than fitting. In order to increase trustworthiness of DL-based solutions and use them as operational tools against fire disasters, it is crucial to interpret and inspect our model's predictions through explainable AI \cite{samek2021explaining}. This can shed light to the limitations of these models and help us refine our models to become more amenable, simpler and with accountable decisions.


\section{Conclusion}\label{sec:conclusion}

In this work, we formulated daily fire danger forecasting as a machine learning problem and produced a harmonized country-wide datacube \cite{prapas_datacube_2021}, from which we extract datasets in four modalities. We implemented some simple, yet effective prototype DL models exploiting the spatial and temporal context of the inputs, which outperformed a baseline RF classifier in terms of AUROC, suggesting that DL can be a valuable tool for wildfire forecasting. 



\begin{ack}
Research funded by the EU H2020 DeepCube 'Explainable AI pipelines for big Copernicus data'.
\end{ack}

{\small
\bibliography{citations}}

\begin{thebibliography}{18}
\providecommand{\natexlab}[1]{#1}
\providecommand{\url}[1]{\texttt{#1}}
\expandafter\ifx\csname urlstyle\endcsname\relax
  \providecommand{\doi}[1]{doi: #1}\else
  \providecommand{\doi}{doi: \begingroup \urlstyle{rm}\Url}\fi

\bibitem[Pettinari and Chuvieco(2020)]{pettinari_fire_2020}
M.~Lucrecia Pettinari and Emilio Chuvieco.
\newblock Fire {Danger} {Observed} from {Space}.
\newblock \emph{Surveys in Geophysics}, 41\penalty0 (6):\penalty0 1437--1459,
  November 2020.
\newblock ISSN 1573-0956.
\newblock \doi{10.1007/s10712-020-09610-8}.
\newblock URL \url{https://doi.org/10.1007/s10712-020-09610-8}.

\bibitem[Turco et~al.(2018)Turco, Rosa-Cánovas, Bedia, Jerez, Montávez,
  Llasat, and Provenzale]{turco_exacerbated_2018}
Marco Turco, Juan~José Rosa-Cánovas, Joaquín Bedia, Sonia Jerez, Juan~Pedro
  Montávez, Maria~Carmen Llasat, and Antonello Provenzale.
\newblock Exacerbated fires in {Mediterranean} {Europe} due to anthropogenic
  warming projected with non-stationary climate-fire models.
\newblock \emph{Nature communications}, 9\penalty0 (1):\penalty0 1--9, 2018.
\newblock Publisher: Nature Publishing Group.

\bibitem[San-Miguel-Ayanz et~al.(2013)San-Miguel-Ayanz, Schulte, Schmuck, and
  Camia]{SANMIGUELAYANZ201319}
Jesús San-Miguel-Ayanz, Ernst Schulte, Guido Schmuck, and Andrea Camia.
\newblock The european forest fire information system in the context of
  environmental policies of the european union.
\newblock \emph{Forest Policy and Economics}, 29:\penalty0 19--25, 2013.
\newblock ISSN 1389-9341.
\newblock \doi{https://doi.org/10.1016/j.forpol.2011.08.012}.

\bibitem[Hantson et~al.(2016)Hantson, Arneth, Harrison, Kelley, Prentice,
  Rabin, Archibald, Mouillot, Arnold, and Artaxo]{hantson_status_2016}
Stijn Hantson, Almut Arneth, Sandy~P. Harrison, Doug~I. Kelley, I.~Colin
  Prentice, Sam~S. Rabin, Sally Archibald, Florent Mouillot, Steve~R. Arnold,
  and Paulo Artaxo.
\newblock The status and challenge of global fire modelling.
\newblock \emph{Biogeosciences}, 13\penalty0 (11):\penalty0 3359--3375, 2016.
\newblock Publisher: Copernicus Publications.

\bibitem[Reichstein et~al.(2019)Reichstein, Camps-Valls, Stevens, Jung,
  Denzler, and Carvalhais]{reichstein_deep_2019}
Markus Reichstein, Gustau Camps-Valls, Bjorn Stevens, Martin Jung, Joachim
  Denzler, and Nuno Carvalhais.
\newblock Deep learning and process understanding for data-driven {Earth}
  system science.
\newblock \emph{Nature}, 566\penalty0 (7743):\penalty0 195--204, 2019.
\newblock Publisher: Nature Publishing Group.

\bibitem[Zhang et~al.(2019)Zhang, Wang, and Liu]{zhang_forest_2019}
Guoli Zhang, Ming Wang, and Kai Liu.
\newblock Forest fire susceptibility modeling using a convolutional neural
  network for yunnan province of china.
\newblock \emph{International Journal of Disaster Risk Science}, 10\penalty0
  (3):\penalty0 386--403, 2019.
\newblock Publisher: Springer.

\bibitem[Huot et~al.(2020)Huot, Hu, Ihme, Wang, Burge, Lu, Hickey, Chen, and
  Anderson]{huot_deep_2020}
Fantine Huot, R.~Lily Hu, Matthias Ihme, Qing Wang, John Burge, Tianjian Lu,
  Jason Hickey, Yi-Fan Chen, and John Anderson.
\newblock Deep {Learning} {Models} for {Predicting} {Wildfires} from
  {Historical} {Remote}-{Sensing} {Data}.
\newblock \emph{arXiv:2010.07445 [cs]}, October 2020.
\newblock URL \url{http://arxiv.org/abs/2010.07445}.
\newblock arXiv: 2010.07445.

\bibitem[Giglio et~al.(2016)Giglio, Schroeder, and
  Justice]{giglio_collection_2016}
Louis Giglio, Wilfrid Schroeder, and Christopher~O. Justice.
\newblock The collection 6 {MODIS} active fire detection algorithm and fire
  products.
\newblock \emph{Remote Sensing of Environment}, 178:\penalty0 31--41, 2016.

\bibitem[Oliveira et~al.(2021)Oliveira, Torgo, and
  Santos~Costa]{oliveira2021evaluation}
Mariana Oliveira, Lu{\'\i}s Torgo, and V{\'\i}tor Santos~Costa.
\newblock Evaluation procedures for forecasting with spatiotemporal data.
\newblock \emph{Mathematics}, 9\penalty0 (6):\penalty0 691, 2021.

\bibitem[Jain et~al.(2020)Jain, Coogan, Subramanian, Crowley, Taylor, and
  Flannigan]{jain_review_2020}
Piyush Jain, Sean~C.P. Coogan, Sriram~Ganapathi Subramanian, Mark Crowley,
  Steve Taylor, and Mike~D. Flannigan.
\newblock A review of machine learning applications in wildfire science and
  management.
\newblock \emph{Environmental Reviews}, 28\penalty0 (4):\penalty0 478--505,
  2020.
\newblock \doi{10.1139/er-2020-0019}.
\newblock URL \url{https://doi.org/10.1139/er-2020-0019}.

\bibitem[Prapas et~al.(2021)Prapas, Kondylatos, and
  Papoutsis]{prapas_datacube_2021}
Ioannis Prapas, Spyros Kondylatos, and Ioannis Papoutsis.
\newblock A datacube for the analysis of wildfires in greece, June 2021.
\newblock URL \url{https://doi.org/10.5281/zenodo.4943354}.

\bibitem[Mu{\~n}oz-Sabater et~al.(2021)Mu{\~n}oz-Sabater, Dutra,
  Agust{\'\i}-Panareda, Albergel, Arduini, Balsamo, Boussetta, Choulga,
  Harrigan, Hersbach, et~al.]{munoz2021era5}
Joaqu{\'\i}n Mu{\~n}oz-Sabater, Emanuel Dutra, Anna Agust{\'\i}-Panareda,
  Cl{\'e}ment Albergel, Gabriele Arduini, Gianpaolo Balsamo, Souhail Boussetta,
  Margarita Choulga, Shaun Harrigan, Hans Hersbach, et~al.
\newblock Era5-land: A state-of-the-art global reanalysis dataset for land
  applications.
\newblock \emph{Earth System Science Data Discussions}, pages 1--50, 2021.

\bibitem[B{\"u}ttner(2014)]{buttner_corine_2014}
Gy{\"o}rgy B{\"u}ttner.
\newblock Corine land cover and land cover change products.
\newblock In \emph{Land use and land cover mapping in Europe}, pages 55--74.
  Springer, 2014.

\bibitem[Bashfield and Keim(2011)]{bashfield_continent-wide_2011}
A.~Bashfield and A.~Keim.
\newblock Continent-wide {DEM} creation for the european union.
\newblock In \emph{34th International Symposium on Remote Sensing of
  Environment. The {GEOSS} Era: Towards Operational Environmental Monitoring.
  Sydney, Australia}, pages 10--15. Citeseer, 2011.

\bibitem[Ager et~al.(2014)Ager, Preisler, Arca, Spano, and
  Salis]{https://doi.org/10.1002/env.2269}
A.~A. Ager, H.~K. Preisler, B.~Arca, D.~Spano, and M.~Salis.
\newblock Wildfire risk estimation in the mediterranean area.
\newblock \emph{Environmetrics}, 25\penalty0 (6):\penalty0 384--396, 2014.
\newblock \doi{https://doi.org/10.1002/env.2269}.
\newblock URL \url{https://onlinelibrary.wiley.com/doi/abs/10.1002/env.2269}.

\bibitem[Shi et~al.(2015)Shi, Chen, Wang, Yeung, Wong, and
  Woo]{shi_convolutional_2015}
Xingjian Shi, Zhourong Chen, Hao Wang, Dit-Yan Yeung, Wai-Kin Wong, and
  Wang-chun Woo.
\newblock Convolutional {LSTM} network: {A} machine learning approach for
  precipitation nowcasting.
\newblock \emph{Advances in neural information processing systems},
  28:\penalty0 802--810, 2015.

\bibitem[Jospin et~al.(2020)Jospin, Buntine, Boussaid, Laga, and
  Bennamoun]{jospin_hands-bayesian_2020}
Laurent~Valentin Jospin, Wray Buntine, Farid Boussaid, Hamid Laga, and Mohammed
  Bennamoun.
\newblock Hands-on {Bayesian} {Neural} {Networks} -- a {Tutorial} for {Deep}
  {Learning} {Users}.
\newblock \emph{arXiv:2007.06823 [cs, stat]}, July 2020.
\newblock URL \url{http://arxiv.org/abs/2007.06823}.
\newblock arXiv: 2007.06823.

\bibitem[Samek et~al.(2021)Samek, Montavon, Lapuschkin, Anders, and
  M{\"u}ller]{samek2021explaining}
Wojciech Samek, Gr{\'e}goire Montavon, Sebastian Lapuschkin, Christopher~J
  Anders, and Klaus-Robert M{\"u}ller.
\newblock Explaining deep neural networks and beyond: A review of methods and
  applications.
\newblock \emph{Proceedings of the IEEE}, 109\penalty0 (3):\penalty0 247--278,
  2021.

\end{thebibliography}

\end{document}